\documentclass[runningheads]{llncs}
\usepackage{graphicx}
\usepackage{tikz}
\usepackage{comment}
\usepackage{amsmath,amssymb} % define this before the line numbering.
\usepackage{color}

\usepackage[accsupp]{axessibility}  % Improves PDF readability for those with disabilities.

\usepackage{amsmath}
\usepackage{subcaption}
\usepackage{booktabs}

\usepackage{xspace}
\usepackage{pgfplots}
\pgfplotsset{compat=newest}
\usepackage{tikz}
\usepackage{pstricks}

\usepackage{siunitx}
\usepackage{microtype}

\usepackage{acronym}

\usepackage[hyphens]{url}
\usepackage[pagebackref=true,breaklinks=true]{hyperref}
\hypersetup{
     colorlinks,
     linkcolor={red!75!black},
     citecolor={blue!75!black},
     urlcolor={blue!75!black},
}
\usepackage[capitalise]{cleveref}

\usepackage[inline]{enumitem}
\setlist*[enumerate]{label=(\arabic*)}

\usepackage{xspace}
\makeatletter
\DeclareRobustCommand\onedot{\futurelet\@let@token\@onedot}
\newcommand{\@onedot}{\ifx\@let@token.\else.\null\fi\xspace}
\makeatother

\newcommand{\etal}{\emph{et~al\onedot}}
\newcommand{\ie}{i.\,e.,\xspace}
\newcommand{\eg}{e.\,g.,\xspace}

\definecolor{bblue}{HTML}{4F81BD}
\definecolor{rred}{HTML}{C0504D}
\definecolor{ggreen}{HTML}{00AA00}
\definecolor{vviolet}{RGB}{128,0,128}
\definecolor{yyellow}{RGB}{255,204,0}
\definecolor{oorange}{RGB}{255,102,0}
\definecolor{ppedrol}{RGB}{0,128,128}
\definecolor{royalblue}{RGB}{0,85,212}
\definecolor{faublue}{RGB}{0,51,102}

\newenvironment{customlegend}[1][]{%
	\begingroup
	\csname pgfplots@init@cleared@structures\endcsname
	\pgfplotsset{#1}%
}{%
	\csname pgfplots@createlegend\endcsname
\endgroup
}%
\def\addlegendimage{\csname pgfplots@addlegendimage\endcsname}

\usepackage{fancyhdr}
\fancypagestyle{specialfooter}{%
  \fancyhf{}
  
  \fancyfoot[L]{\scriptsize{$\copyright$ Vision for Art (VISART) Workshop at the European Conference on Computer Vision (ECCV) 2022, scheduled for October 23, 2022 in Tel Aviv.}}
}
\begin{document}
% \renewcommand\thelinenumber{\color[rgb]{0.2,0.5,0.8}\normalfont\sffamily\scriptsize\arabic{linenumber}\color[rgb]{0,0,0}}
% \renewcommand\makeLineNumber {\hss\thelinenumber\ \hspace{6mm} \rlap{\hskip\textwidth\ \hspace{6.5mm}\thelinenumber}}
% \linenumbers
\pagestyle{headings}
\mainmatter
\def\ECCVSubNumber{31}  % Insert your submission number here

\title{ArtFacePoints: High-resolution Facial Landmark Detection in Paintings and Prints}

% INITIAL SUBMISSION 
\begin{comment}
\titlerunning{ECCV-22 submission ID \ECCVSubNumber} 
\authorrunning{ECCV-22 submission ID \ECCVSubNumber} 
\author{Anonymous ECCV submission}
\institute{Paper ID \ECCVSubNumber}
\end{comment}
%******************

% CAMERA READY SUBMISSION
%\begin{comment}
\titlerunning{ArtFacePoints} %: High-resolution Facial Landmark Detection}
% If the paper title is too long for the running head, you can set
% an abbreviated paper title here
%
\author{Aline Sindel \and
Andreas Maier \and
Vincent Christlein}
\authorrunning{A. Sindel et al.}
% First names are abbreviated in the running head.
% If there are more than two authors, 'et al.' is used.
%
\institute{Pattern Recognition Lab, FAU Erlangen-N\"urnberg, Germany\\
\email{aline.sindel@fau.de}}
%\end{comment}
%******************
\maketitle
%
% Add copyright message in footer line
\thispagestyle{specialfooter}

\begin{abstract}
Facial landmark detection plays an important role for the similarity analysis in artworks to compare portraits of the same or similar artists. With facial landmarks, portraits of different genres, such as paintings and prints, can be automatically aligned using control-point-based image registration. We propose a deep-learning-based method for facial landmark detection in high-resolution images of paintings and prints. It divides the task into a global network for coarse landmark prediction and multiple region networks for precise landmark refinement in regions of the eyes, nose, and mouth that are automatically determined based on the predicted global landmark coordinates. We created a synthetically augmented facial landmark art dataset including artistic style transfer and geometric landmark shifts. Our method demonstrates an accurate detection of the inner facial landmarks for our high-resolution dataset of artworks while being comparable for a public low-resolution artwork dataset in comparison to competing methods.
\end{abstract}

\section{Introduction}
Facial landmark detection is a key element to analyze face characteristics. In art investigations, portraits of the same artist but also portraits of artists with a similar style are analyzed by comparing facial structures, \eg by creating hand-drawn tracings of the facial inner lines or the face outline and manually compare the tracings. In this procedure, the art technologists can be supported by automatically detecting facial landmarks in the portrait images and using them to align the images for comparison. 
Facial landmark detection is a widely explored field for natural images, but is not directly applicable to artworks, since artworks show larger variation in the texture of the images and in the geometry of the landmarks than natural images. In art technology, art examinations are usually based on high-resolution or even macrophotographic images of the artwork. 
Therefore, there is a demand for a method that can accurately detect facial landmarks in high-resolution images of artworks.

\begin{figure}[t]
\centering
\includegraphics[width=\textwidth]{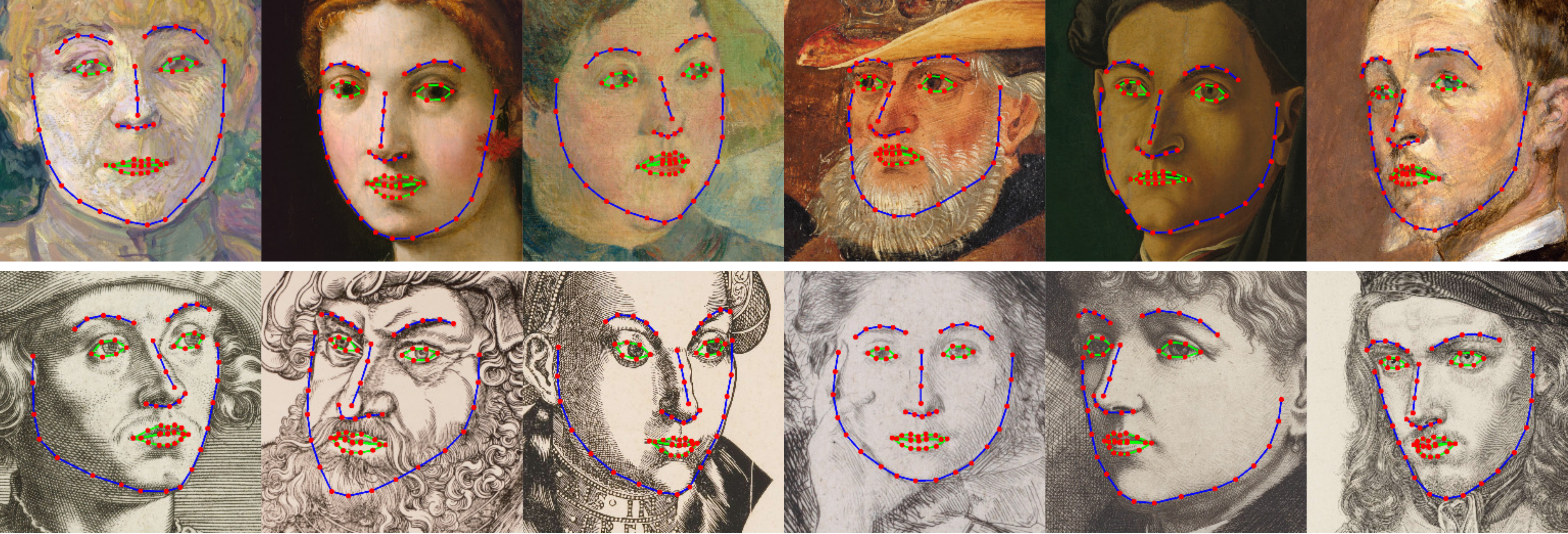}
\caption{Our ArtFacePoints accurately detects facial landmarks in high-resolution paintings and prints. \\ \\
\tiny{Image sources: Details of (a) Henri de Toulouse-Lautrec, The Streetwalker, The Metropolitan Museum of Art, 2003.20.13;
(b) Florentine 16th Century (painter), Portrait of a Young Woman, Widener Collection, National Gallery of Art, 1942.9.51;
(c) Paul Gauguin, Madame Alexandre Kohler, Chester Dale Collection, National Gallery of Art, 1963.10.27;
(d) Meister des Augustiner-Altars (Hans Traut) und Werkstatt mit Rueland Frueauf d.Ä., Marter der Zehntausend, Germanisches Nationalmuseum Nürnberg, on loan from Museen der Stadt Nürnberg, Kunstsammlungen, Gm 149;
(e) North Italian 15th Century (painter), Portrait of a Man, Samuel H. Kress Collection, National Gallery of Art, 1939.1.357;
(f) Henri de Toulouse-Lautrec, Albert (René) Grenier, The Metropolitan Museum of Art, 1979.135.14;
(g) Cornelis Cort (after Albrecht Dürer), Portret van Barent van Orley, Rijksmuseum, RP-P-2000-140;
(h) Lucas Cranach the Elder, Johann I. der Beständige, Kurfürst von Sachsen, Staatliche Kunstsammlungen Dresden SKD, Kupferstich-Kabinett, A 129786, SLUB / Deutsche Fotothek, Loos, Hans, \url{http://www.deutschefotothek.de/documents/obj/70243984} (Free access - rights reserved);
(i) Hans Brosamer, Sybille von Cleve, Herzog Anton Ulrich-Museum, HBrosamer AB 3.33H;
(j) Edgar Degas, Marguerite De Gas the Artist's Sister, The Metropolitan Museum of Art, 2020.10;
(k) Thomas Nast, Portrait of the Artist's Wife Sarah Edwards Nast, Reba and Dave Williams Collection, National Gallery of Art, 2008.115.3695;
(l) Samuel Amsler, Portret Carl Philip Fohr, Rijksmuseum, RP-P-1954-345
}
}
\label{fig-01}
\end{figure} 

In this paper, we propose ArtFacePoints\footnote{Our source code is available at \url{https://github.com/asindel/ArtFacePoints}}, a deep-learning-based facial landmark detector for high-resolution images of paintings and prints. It consists of one global and multiple region networks for coarse and fine facial landmark prediction. We employ a ResNet-based encoder-decoder to predict global facial landmarks in downsized images, which are then used to crop regions of the high-resolution image and the global feature maps. With the global feature map regions as prior information, the region networks get an additional impetus to refine the global feature maps using the high-resolution image details. 
To train and evaluate our method, we created a large facial landmark art dataset by applying style transfer techniques and geometric facial landmark distortion to a public landmark dataset of natural images and by collecting and annotating images of real artworks.
In \cref{fig-01}, we show some qualitative examples of our ArtFacePoints for different artistic styles.

\section{Related Work}
Facial landmark detection is an active research area for natural images~\cite{WuY2019}, hence we will only summarize a selection of regression-based approaches. 
For instance, one example for a machine-learning-based regression facial landmark method is dlib~\cite{KazemiV2014} which employs an ensemble of regression trees.
Using convolutional neural networks (CNN) we can broadly differ between methods that directly regress the coordinates using a fully connected layer~\cite{SunY2013,ZhangZ2014,HeKe2016}, or direct heatmap-based approaches~\cite{SunK2019,BulatA2017,KowalskiM2017,WangX2019} that draw the coordinates as Gaussian peaks and formulate the task as the regression between the predicted and target heatmaps, and indirect heatmap-based approaches~\cite{HonariS2018,RobinsonJP2019,ChandranP2020,KordonF2021} that use the differentiable spatial softargmax~\cite{NibaliA2018} to extract the coordinates from the heatmaps instead of using the argmax and then compute a loss between the predicted and target coordinates.

The high-resolution network (HR-Net)~\cite{SunK2019_HPE,SunK2019} was developed for different vision tasks such as for human pose estimation and facial landmark detection. It uses multi-resolution blocks which connect high-to-low resolution convolutions to maintain high-resolution representations. The coordinates of the landmarks are extracted from the heatmap based on the highest response peaks. Recently, the HR-Net was also used for human pose estimation in Greek vase paintings~\cite{MadhuP2020}, for which style transfer using adaptive instance normalization~\cite{HuangX2017} was applied to a labeled dataset of natural images to train the model.

Chandran \etal~\cite{ChandranP2020} proposed a region based facial landmark detector for high-resolution images up to 4K. They detect the coarse landmark positions in a downsized image using the heatmap-based Hourglass network and softargmax~\cite{NibaliA2018}. By using attention-driven cropping based on the global landmark locations, they extract regions around the eyes, nose, and mouth in the high-resolution image. They predict more accurate landmarks in the regions using for each region specifically trained Hourglass networks. In our approach, we are strongly oriented towards their idea of high-resolution landmarks with a global and regional detection network, however, we make some architectural and conceptual changes, such as using a ResNet-based encoder-decoder or the use of a global feature map as additional input for the regional networks.

For facial landmark detection in artworks, Yaniv \etal~\cite{YanivJ2019} proposed Face of Art, which predicts the landmarks in three steps. First, the global landmark location is estimated from a response map, then the landmarks are corrected using a pretrained point distribution model and finally they are tuned using weighted regularized mean shift. They created a synthetic dataset using the style transfer method of Gatys \etal~\cite{GatysL2015} for low-resolution images and geometric landmark augmentations. The latter we adopt for our work.
The facial landmark detection method for mangas~\cite{StrickerM2018} uses the deep alignment network~\cite{KowalskiM2017} which uses multiple steps to refine the heatmaps of the previous steps.

\begin{figure}[t]
	\centering
\begin{subfigure}[b]{\linewidth}
\includegraphics[width=\textwidth]{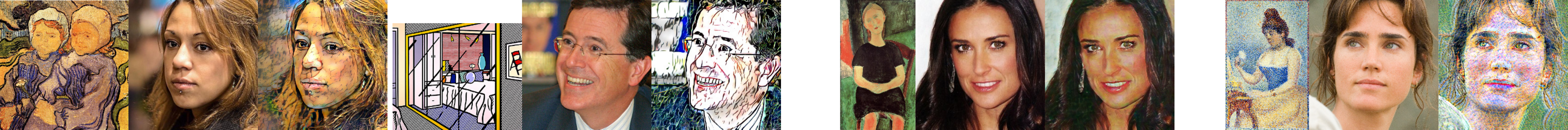}
\caption{~AdaIN Synthetic Paintings (Style, content, and generated image)}\label{fig-02:syn_adain_paint}
\end{subfigure}
\begin{subfigure}[b]{\linewidth}
\includegraphics[width=\textwidth]{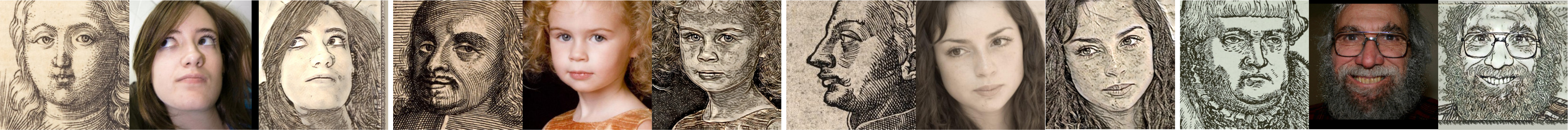}
\caption{~AdaIN Synthetic Prints (Style, content, and generated image)}\label{fig-02:syn_adain_print}
\end{subfigure}
\begin{subfigure}[b]{.49\linewidth}
\includegraphics[width=\textwidth]{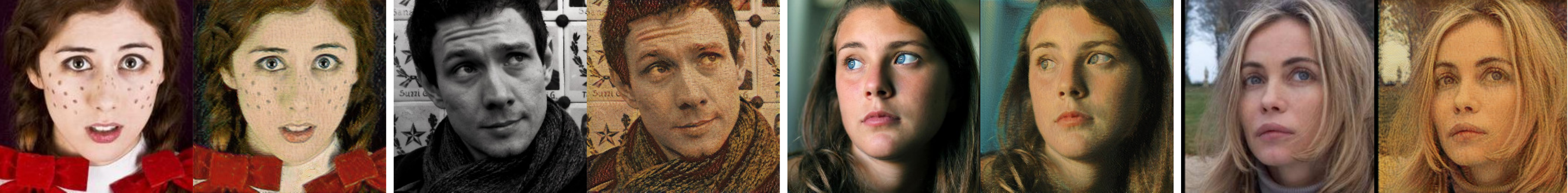}
\caption{~CycleGAN Synthetic Paintings}\label{fig-02:syn_cyclegan_paint}
\end{subfigure}
\begin{subfigure}[b]{.49\linewidth}
\includegraphics[width=\textwidth]{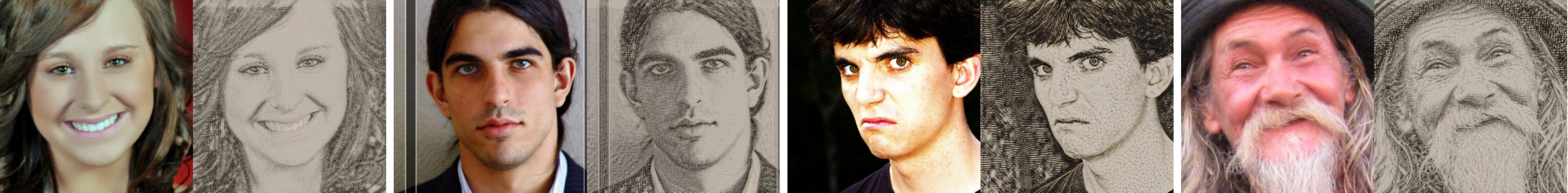}
\caption{~CycleGAN Synthetic Prints}\label{fig-02:syn_cyclegan_print}
\end{subfigure}
	\caption{High-resolution synthetic image generation using AdaIN and CycleGAN. \\ \\
\tiny{Image sources: (a) Paintings: 
Vincent van Gogh, Two Children, Musée d'Orsay, Paris, WikiArt;
Roy Lichtenstein, Interior with Mirrored Closet, 1991, $\copyright$ Estate of Roy Lichtenstein;
Amedeo Modigliani, Seated Young Woman, Private Collection, WikiArt;
Georges Seurat, Study for Young Woman Powdering Herself, WikiArt;
(b) Prints: 
Albrecht Dürer, The Virgin and Child Crowned by One Angel (Detail), Rosenwald Collection, National Gallery of Art, 1943.3.3546;
Unknown, Adam Zusner (Detail). Wellcome Collection, 9833i;
T. Stimmer, Paolo Giovio (Detail), Wellcome Collection, 3587i;
Lucas Cranach the Younger, Sigmund (Halle, Erzbischof) (Detail), Staatliche Kunstsammlungen Dresden, Kupferstich-Kabinett, Nr.: B 376, 1, SLUB / Deutsche Fotothek, Kramer, Rudolph, \url{https://www.deutschefotothek.de/documents/obj/70231557} (Free access - rights reserved);
(a-d) Photos: Images from the 300-W dataset
}	
	}
	\label{fig-02}
\end{figure}

\section{Artistic Facial Landmarks Dataset Creation} \label{DatasetCreation}

For the training and testing of our method, we collected high-resolution portrait images of different artistic styles from various museums and institutes such as from the Cranach Digital Archive (CDA), Germanisches Nationnalmuseum Nürnberg (GNM), The Metropolitan Museum of Art (MET), National Gallery of Art (NGA), and Rijksmuseum, and additionally also selected images from the WikiArt dataset~\cite{NicholK2016} for training. In order to automatically detect and crop the face region in the images of paintings and prints, we trained and applied the object detector YOLOv4~\cite{BochkovskiyA2020}. The cropped images are either used as surrogates to generate synthetic artworks or are manually labeled.

\subsection{Synthetic image generation using style transfer and geometric augmentations}
There exist multiple public datasets for facial landmark detection with ground truth labels such as 300-W~\cite{SagonasC2013}, however not yet for high-resolution paintings and prints. Thus, we apply style transfer and image-to-image translation techniques to the 300-W dataset to transform the images scaled to $1024 \times 1024 \times 3$ pixels into synthetic paintings and prints.

As style transfer method, we use adaptive instance normalization (AdaIN)~\cite{HuangX2017} which aligns the channel-wise mean and variance of the content image with the style image. AdaIN was trained using MS-COCO~\cite{LinTY2014} for the content and WikiArt images~\cite{NicholK2016} for the style images. For our application, we set the alpha parameter, which controls the trade-off between content and style, to $1$ to achieve full stylization and apply both content and style image in $1024 \times 1024$ resolution. Some qualitative results of our high-resolution synthetically stylized portrait images are shown in~\cref{fig-02:syn_adain_paint,fig-02:syn_adain_print}. The synthetic paintings and prints reuse the textures and the color distribution from the style images resulting in a motley set of images.

Further, we use the unpaired image-to-image translation technique CycleGAN~\cite{ZhuJY2017} to learn a mapping between faces in photographs and artworks by exploiting cycle consistency. 
We train one CycleGAN for each domain pair, \ie one for photo-to-print and one for photo-to-painting in lower resolution $512 \times 512 \times 3$ and apply them in high-resolution (see~\cref{fig-02:syn_cyclegan_paint,fig-02:syn_cyclegan_print}).  
Interestingly, the synthetic paintings using CycleGAN clearly show crack structures (craquelure) in the paint typical for old paintings. 
The synthetic prints express the shading and continuous regions of the photos with tiny lines to mimic the real prints.

\begin{figure}[t]
\centering
\includegraphics[width=\textwidth]{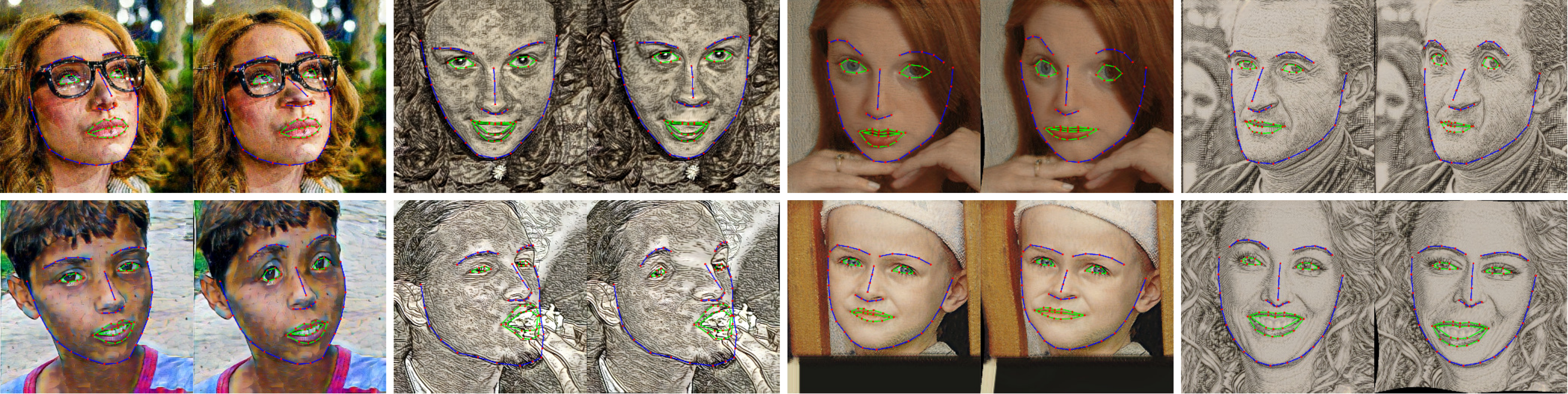}
\caption{Geometric augmentations of ground truth facial landmarks in synthetic paintings and prints: before and after applying the geometric transformation.}
\label{fig-03}
\end{figure}

Artistic faces not only differ to photos in their textual style but also in their geometric arrangement of the facial landmarks. Faces can be longitudinally or horizontally stretched or unbalanced with, \eg larger eyes. To account for some degree of artistic variations, we adopt the geometric augmentation strategy of Yaniv et al.~\cite{YanivJ2019} to randomly shift or resize single groups of landmarks such as the eyes or mouth, or stretch or squeeze the face. Based on the movements of the landmarks a thin-plate-spline displacement field is computed which is used to warp the synthetic artwork~\cite{YanivJ2019}. Some visual examples in \cref{fig-03} show synthetic images with ground truth annotations in their original pose and after geometric warping of the images.

\subsection{Semi-automatic facial landmarks annotation}
Additionally to the synthetic dataset, we also include a small number of real artworks for training and validation. For those and also for our test dataset, we annotated the facial landmarks in a semi-automatic manner. 
We applied the random forest based facial landmark detector dlib~\cite{KazemiV2014} to the images and manually corrected the landmarks or in case of a dlib failure, we annotated the landmarks from scratch. For performing the annotations, we have written a small graphical user interface tool in Python that allows annotating from scratch and annotation refinement by enabling the user to move the single landmarks via mouse drag of the items.

\begin{figure}[t]
\centering
\includegraphics[width=\textwidth]{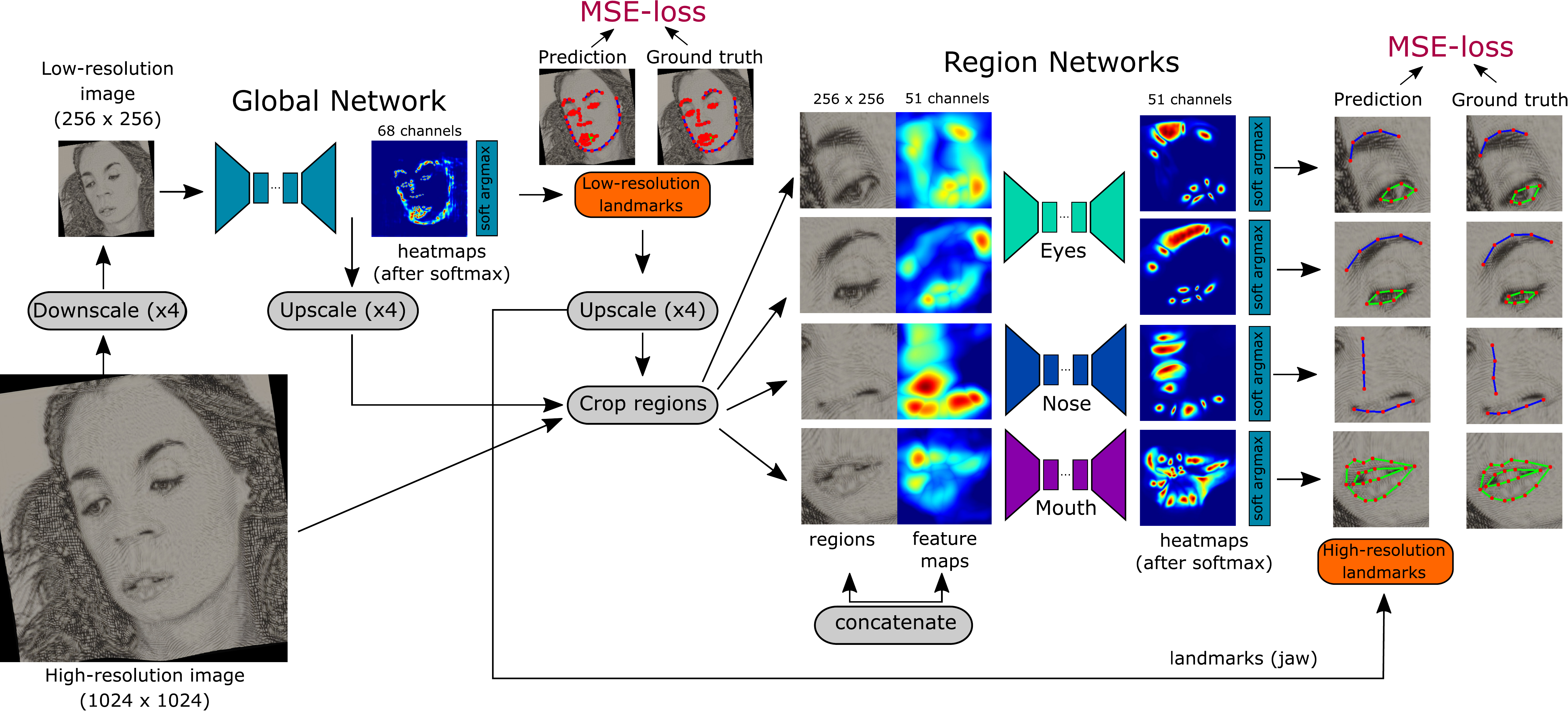}
\caption{Our facial landmarks detector ArtFacePoints for high-resolution images of paintings and prints splits the task into a global coarse prediction step and a regional refinement step. The high-resolution image is downsized and fed to the global ResNet encoder-decoder network, which predicts the low-resolution landmarks using softmax on a 68 channels heatmap. Based on the low-resolution landmarks, regions of eyes, nose, and mouth are extracted from the high-resolution image and the upscaled global feature map. For each region, the image patch and feature maps patches are concatenated and are fed to its specific region network to predict the high-resolution landmarks.
The global as well as the regional landmarks are compared to the ground truth landmarks using the mean squared error (MSE) loss.}
\label{fig-04}
\end{figure}

\section{ArtFacePoints}
In this section, our method for high-resolution facial landmark detection in artworks is described, which consists of two main steps, the global and the regional facial landmark detection, as illustrated in \cref{fig-04}.

\subsection{Global facial landmark detection}
For the global network, we use the encoder-decoder network with stacked ResNet~\cite{HeK2016} blocks in the bottleneck, which was originally employed for style transfer~\cite{JohnsonJ2016}. We exchanged the transpose convolutional layers in the decoder with a $3 \times 3$ convolutional layer and bicubic upsampling to obtain smoother heatmaps. The input to the network is the low-resolution RGB image and the prediction are $N_G = 68$ heatmaps with the same width and height as the input image.
Each heatmap should mark the location of its respective facial landmark as a peak with the highest value. Then, we use the spatial softargmax~\cite{NibaliA2018} operator to extract the landmarks from the heatmaps in a differentiable manner.

\begin{table}[t]
\centering
\caption{Style Art Faces Dataset: Number of images including facial landmarks (68 per image) for the real and the synthetic artworks.} 
\label{tab-01}
\begin{tabular*}{\textwidth}{l@{\extracolsep\fill}rrrrrrr}
\toprule
Dataset  & Real & Real & AdaIN  & AdaIN  & CycleGAN  & CycleGAN  & Total\\
  & Paintings & Prints & Paintings &  Prints &  Paintings &  Prints & \\
\midrule
Train & 160 & 160 & 511 & 511 & 511 & 511 & \textbf{2361} \\ 
Val & 30 & 30 & 220 & 220 & 220 & 220 & \textbf{940} \\
Test & 40 & 40 & & & & & \textbf{80} \\
\bottomrule
\end{tabular*}
\end{table}

\subsection{Regional facial landmarks refinement}
The global landmark predictions in the low-resolution image are upscaled by a factor of 4 to match the high-resolution image. Similar to~\cite{ChandranP2020}, we automatically extract regions in the high-resolution image around the landmark predictions of the nose, the mouth, and each eye including the eye brow. The region size is padded with a random value between $0.25$ to $0.5$ of the original region size that is directly estimated from the global landmarks. For inference, the padding is a fixed value of $0.25$ of the region size. 
Analogously, we also extract the same regions from the upscaled feature maps of the global network, which is the direct output of the global network before applying spatial softmax. All regions are scaled to a fixed patch size. The three region networks for eye, nose, and mouth also use the ResNet-based encoder-decoder architecture like the global network, but instead of only feeding the RGB patch, we concatenate the RGB patch ($3$ channels) with the corresponding regions of the feature maps ($N_r$ channels) as input, where $N_r$ depends on the specific region network (eye: $11$, nose: $9$, mouth: $20$). 
With the channel fusion, the region network gets the global location of the landmark as prior information, which supports the refinement task. 
There is no weight sharing between the region networks, such that each network can learn its specific features for the facial sub regions. The high-resolution landmarks are also extracted using spatial softargmax~\cite{NibaliA2018}.

We use the mean squared error (MSE) loss of predicted landmarks and ground truth landmarks for both the global and regional landmark detection task:
\begin{equation}
\mathcal{L}_{\text{MSE}} = \frac{1}{N_G} \sum_i^{N_G} ( \hat{\mathbf{x}}_i - \mathbf{x}_i )^{2} + \lambda \sum_r^{4} \frac{1}{N_r} \sum_j^{N_r} ( \hat{\mathbf{y}}_j - \mathbf{y}_j )^{2},
\end{equation}
where $\lambda$ is a weighting factor, $\hat{\mathbf{x}}$, $\mathbf{x}$ are the predicted and ground truth global coordinates and $\hat{\mathbf{y}}$, $\mathbf{y}$ are the predicted and ground truth coordinates of one region.

%\subsection{Fusion of global and regional facial landmarks}
For inference, we need to transfer the high-resolution landmarks from the individual regions back to the global coordinate system. Therefore, we track both, the bounding box coordinates of the extracted regions and the original region sizes. Then, we scale the local coordinates by the region size and add the offset of the bounding box. For the jaw line, we only have the global estimate. Hence, the complete facial landmark prediction is the combination of the global jaw line and the refined regions.

\begin{table}[t]
\centering
\caption{Quantitative results for our paintings and prints test dataset ($1024 \times 1024$ using the mean error of the 68 landmarks and 51 high-resolution landmarks (without jaw line). * For dlib only 38 out of 40 paintings and 34 out of 40 prints were detected.} 
\label{tab-02}
\scriptsize
%\tiny
\begin{tabular*}{\textwidth}{l@{\extracolsep\fill}rrrr}
\toprule
Metrics & \multicolumn{2}{c}{Paintings} &  \multicolumn{2}{c}{Prints}  \\
Mean Error (ME) & 68 landmarks & 51 landmarks & 68 landmarks & 51 landmarks\\ 
\midrule
dlib* & 20.03$\pm$7.77 & 17.53$\pm$7.25 & 43.18$\pm$20.99 & 34.94$\pm$20.02 \\ 
dlib* (Art) & 33.39$\pm$24.49 & 26.84$\pm$24.11 & 109.09$\pm$99.81 & 104.12$\pm$108.01 \\ 
HR-Net & 22.29$\pm$7.24 & 19.22$\pm$4.36 & 35.90$\pm$16.87 & 26.28$\pm$16.23 \\ 
HR-Net (Art) & 20.01$\pm$6.99 & 17.05$\pm$3.50 & 27.81$\pm$9.00 & 19.67$\pm$4.44 \\ 
Face of Art & \textbf{17.93}$\pm$7.76 & 14.33$\pm$4.81 & 27.29$\pm$8.44 & 18.89$\pm$5.94 \\ 
ArtFacePoints (global) & 18.87$\pm$9.23 & 13.42$\pm$4.80 & 26.37$\pm$8.82 & 16.74$\pm$5.22 \\ 
ArtFacePoints (w/o FM) & 17.97$\pm$8.51 & 12.62$\pm$4.62 & \textbf{25.60}$\pm$8.50 & 15.83$\pm$4.79 \\ 
ArtFacePoints & 18.88$\pm$8.52 & \textbf{12.45}$\pm$4.44 & 25.65$\pm$8.83 & \textbf{15.78}$\pm$6.20 \\ 
\bottomrule
\end{tabular*}
\end{table}

\section{Experiments and Results}

\subsection{Datasets}
Our high-resolution facial landmarks dataset of artworks is comprised of real and synthetic paintings and prints as described in \cref{DatasetCreation}. 
The number of images for training, validation, and test split are summarized in \cref{tab-01}. All images are of size $1024 \times 1024 \times 3$ and contain each 68 facial landmarks according to the 300-W annotation concept. For the 2924 synthetic artworks, we directly reused the 300-W annotations and for the 460 real artworks, we semi-automatically annotated the landmarks. 

Further, we use the public Artistic Faces Dataset~\cite{YanivJ2019} for comparison. 
It consists of a total of 160 images of size $256 \times 256 \times 3$, which are composed of $10$ images per artist representing different artistic styles. 

\subsection{Implementation and experimental details}
We pretrain the global network for 60 epochs using the Adam optimizer, a learning rate of $\eta=1\cdot10^{-4}$ with linear decay of $\eta$ to $0$ starting at epoch $30$, a batch size of $16$ and early stopping. Then, we initialize the region networks with the weights of the global network and train both jointly for $30$ epochs using $\eta=1\cdot10^{-4}$ with linear decay of $\eta$ to $0$ starting at epoch $10$, a batch size of $4$ and early stopping. The input image size to crop the regions is $1024 \times 1024$ and the patch size for the global and region networks is $256 \times 256$. For the loss computation, both global and local landmarks are normalized to $\left[-0.5, 0.5\right]$ based on their image or region size and $\lambda = 0.25$ is set to weight each region term equally in the loss.

As comparison methods, we use dlib~\cite{KazemiV2014}, HR-Net~\cite{SunK2019}, and Face of Art~\cite{YanivJ2019}. We retrained dlib and fine-tuned HR-Net using our synthetically augmented art facial landmark dataset including the geometric transformations of the landmarks. To measure the performance, we compute the mean Euclidean error (ME) of the predicted and manually labeled facial landmarks. In contrast to related works~\cite{YanivJ2019,SunK2019}, we do not normalize the error based on the inter-ocular distance, inter-pupil distance, or the diagonal of the bounding box. Especially to assess the accuracy in high-resolution images, we prefer to compare the directly measured pixel distance between the landmarks as the denominator using the eye distance or bounding box distance can become very large and thus the error would become very small. 

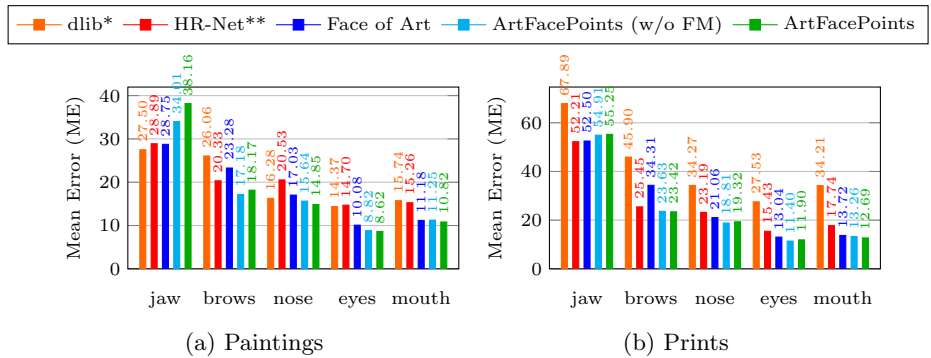
\begin{figure}[t]
\scriptsize
\centering
\begin{tikzpicture}
	\begin{customlegend}[
			legend entries={dlib*,HR-Net**,Face of Art,ArtFacePoints (w/o FM),ArtFacePoints},
			legend style={
				/tikz/every even column/.append	style={column sep=.05cm}},
			legend columns=-1,
			legend cell align=left]
		\addlegendimage{oorange,mark=square*}
		\addlegendimage{red,mark=square*}		
		\addlegendimage{blue,mark=square*}
		\addlegendimage{cyan,mark=square*}
		\addlegendimage{ggreen,mark=square*}				
	\end{customlegend}
\end{tikzpicture}

\subcaptionbox{Paintings\label{fig-05:facial_parts_HR_paint}}{
\begin{tikzpicture}
\begin{axis} [
        width  = 0.25*\textwidth,
        height = 4cm,
        major x tick style = transparent,
        ybar,
        bar width=0.08cm,
    	x=.85cm, 
    	enlarge x limits={abs=0.5cm},    	
        ymajorgrids = true,
        ylabel = {Mean Error (ME)},
      	ymin=0,
        symbolic x coords= {jaw, brows, nose, eyes, mouth}, 
        xtick = data,
        xticklabel style={text height=2ex},
        nodes near coords={\pgfmathprintnumber[fixed zerofill,precision=2]\pgfplotspointmeta},
        every node near coord/.append style={font=\tiny, xshift = 0 , rotate=90, anchor=west},
        nodes near coords align = {center},
        ]
        \addplot [style={oorange,fill=oorange,mark=none}]
        table[x=Region, y=Dlib] 
        {tables/HRArtFaces_partsPaintings.dat};       
        \addplot [style={red,fill=red,mark=none}]
        table[x=Region, y=HRNetArt] 
        {tables/HRArtFaces_partsPaintings.dat};       
        \addplot [style={blue,fill=blue,mark=none}]
        table[x=Region, y=FacesOfArt] 
        {tables/HRArtFaces_partsPaintings.dat};
		\addplot [style={cyan,fill=cyan,mark=none}]
        table[x=Region, y=ArtFacePointswocat] 
        {tables/HRArtFaces_partsPaintings.dat};         
        \addplot [style={ggreen,fill=ggreen,mark=none}]       
        table[x=Region, y=ArtFacePoints] 
        {tables/HRArtFaces_partsPaintings.dat};     
\end{axis}
\end{tikzpicture}
}
\subcaptionbox{Prints\label{fig-05:facial_parts_HR_print}}{
\begin{tikzpicture}
\begin{axis} [
        width  = 0.25*\textwidth,
        height = 4cm,
        major x tick style = transparent,
        ybar,
        bar width=0.08cm,
    	x=.85cm, 
    	enlarge x limits={abs=0.5cm},
        ymajorgrids = true,
        ylabel = {Mean Error (ME)},
      	ymin=0,
        symbolic x coords= {jaw, brows, nose, eyes,  mouth},
        xtick = data,
        xticklabel style={text height=2ex},
        nodes near coords={\pgfmathprintnumber[fixed zerofill,precision=2]\pgfplotspointmeta},
        every node near coord/.append style={font=\tiny, xshift = 0 , rotate=90, anchor=west},
        nodes near coords align = {center},
        ]
        \addplot [style={oorange,fill=oorange,mark=none}]
        table[x=Region, y=Dlib] 
        {tables/HRArtFaces_partsPrints.dat};       
        \addplot [style={red,fill=red,mark=none}]
        table[x=Region, y=HRNetArt] 
        {tables/HRArtFaces_partsPrints.dat};       
        \addplot [style={blue,fill=blue,mark=none}]
        table[x=Region, y=FacesOfArt] 
        {tables/HRArtFaces_partsPrints.dat};
		\addplot [style={cyan,fill=cyan,mark=none}]
        table[x=Region, y=ArtFacePointswocat] 
        {tables/HRArtFaces_partsPrints.dat};         
        \addplot [style={ggreen,fill=ggreen,mark=none}]       
        table[x=Region, y=ArtFacePoints] 
        {tables/HRArtFaces_partsPrints.dat};    
\end{axis}
\end{tikzpicture}
}
\caption{Quantitative comparison of the facial landmark prediction for individual parts of the face of our high-resolution paintings and prints. * For dlib only 38 out of 40 paintings and 34 out of 40 prints were detected. ** HR-Net is the fine-tuned model on our art dataset.} 
\label{fig-05}
\end{figure}

\begin{figure}[ht!]
\centering
\includegraphics[width=\textwidth]{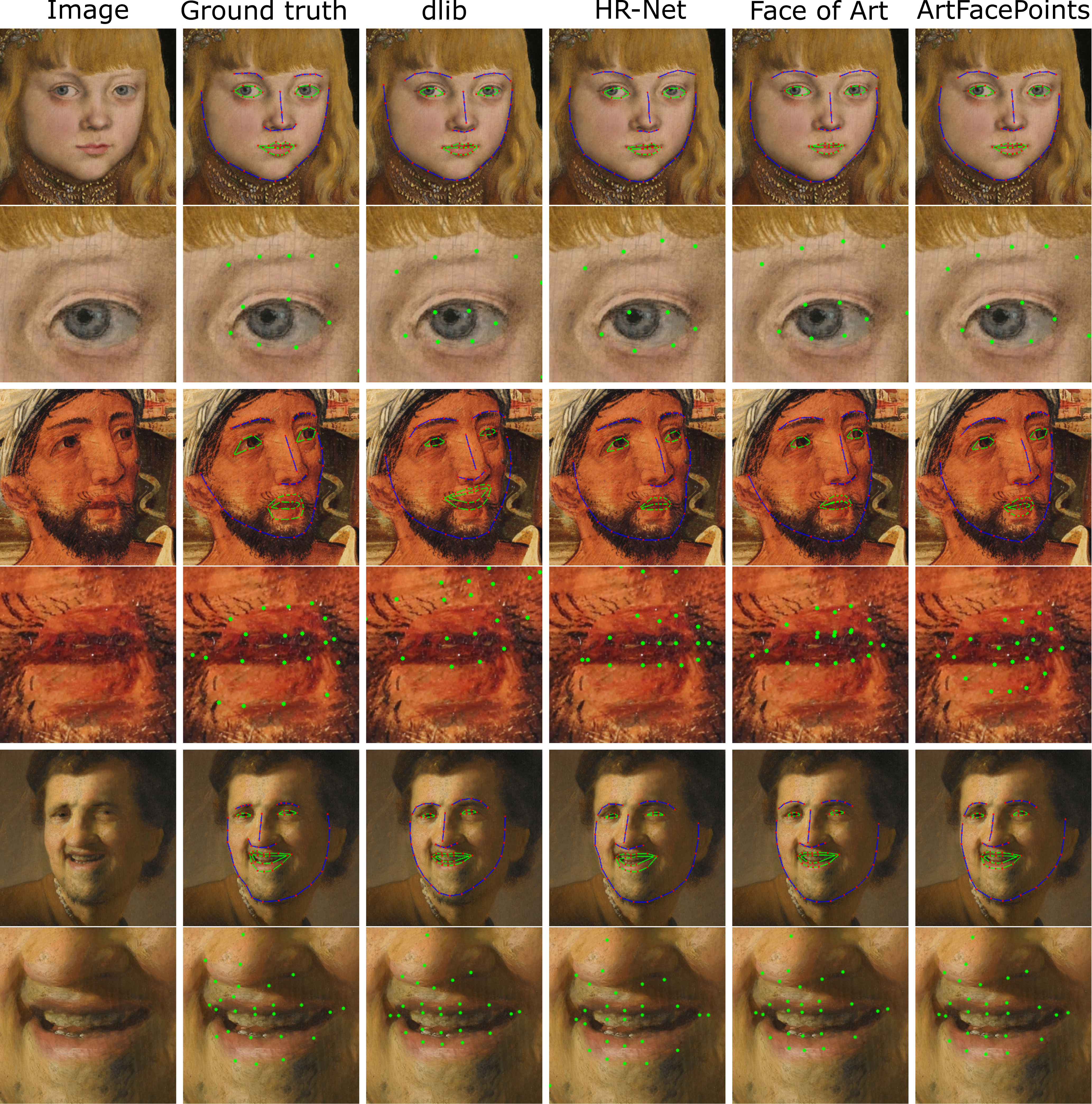}
\caption{Qualitative comparison of facial landmark detection for paintings. \\ \\
\tiny{Image sources: Details of (a) Lucas Cranach the Elder, A Prince of Saxony, Ralph and Mary Booth Collection, National Gallery of Art, 1947.6.1;
(b) Meister des Augustiner-Altars (Hans Traut) und Werkstatt mit Rueland Frueauf d.Ä., Marter der Zehntausend, Germanisches Nationalmuseum Nürnberg, on loan from Museen der Stadt Nürnberg, Kunstsammlungen, Gm 149;
(c) Rembrandt van Rijn (circle of), Laughing Young Man, Rijksmuseum, SK-A-3934
}
}
\label{fig-06}
\end{figure} 

\begin{figure}[t]
\centering
\includegraphics[width=\textwidth]{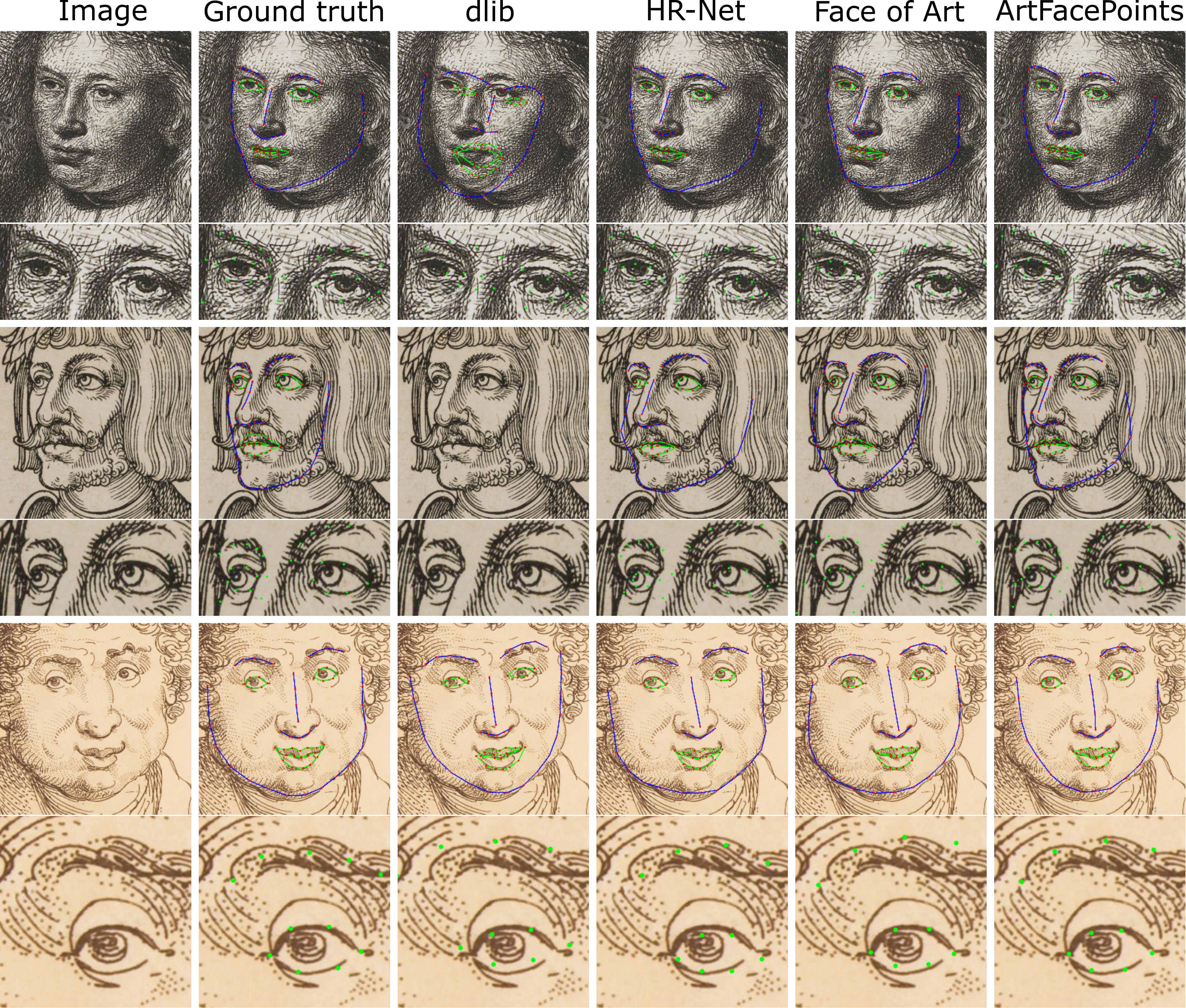}
\caption{Qualitative comparison of facial landmark detection for prints. \\ \\
\tiny{Image sources: Details of (a) Rembrandt van Rijn, The great Jewish bride, Rijksmuseum, RP-P-OB-724;
(b) after Hans Baldung Grien, Ulrich von Hutten, British Museum, London, 1911,0708.39, Photo $\copyright$ Thomas Klinke, courtesy of the Trustees of the British Museum;
(c) Thomas Rowlandson, Joy with Tranquility, The Elisha Whittelsey Collection, The Metropolitan Museum of Art, 59.533.699
}
}
\label{fig-07}
\end{figure} 

\begin{table}[t]
\centering
\caption{Quantitative comparison for the public Artistic Faces dataset ($256 \times 256$). * For dlib only 134 out of 160 artworks were detected.} 
\label{tab-03}
\tiny
\begin{tabular*}{\textwidth}{l@{\extracolsep\fill}rr rrrrr}
\toprule
Mean Error (ME) & 68  & 51 & jaw & eye brows & nose & eyes & mouth\\ 
\midrule
dlib* & 5.75$\pm$3.64 & 5.42$\pm$3.43 & \textbf{6.74}$\pm$5.37 & 7.01$\pm$6.54 & 4.85$\pm$4.10 & 5.54$\pm$3.90 & 4.82$\pm$4.23 \\ 
HR-Net (Art) & 5.28$\pm$2.39 & 4.60$\pm$2.45 & \textbf{7.31}$\pm$3.74 & 5.76$\pm$4.08 & 4.61$\pm$3.58 & 4.19$\pm$3.87 & 4.27$\pm$2.11 \\ 
Face of Art & \textbf{4.87}$\pm$2.48 & \textbf{3.88}$\pm$2.36 & 7.83$\pm$4.44 & 5.57$\pm$6.02 & 3.90$\pm$4.82 & 3.23$\pm$2.54 & \textbf{3.41}$\pm$3.06 \\ 
ArtFacePoints (w/o FM) & 5.77$\pm$3.12 & 4.27$\pm$2.66 & 10.29$\pm$6.37 & 5.83$\pm$5.24 & 4.51$\pm$3.29 & \textbf{3.19}$\pm$3.48 & 4.02$\pm$3.17 \\ 
ArtFacePoints & 5.58$\pm$3.19 & 3.91$\pm$2.77 & 10.60$\pm$6.08 & \textbf{5.55}$\pm$6.32 & \textbf{3.77}$\pm$2.57 & 3.24$\pm$3.88 & 3.55$\pm$2.97 \\ 
\bottomrule
\end{tabular*}
\end{table}

\subsection{Results}
The quantitative results for our high-resolution art dataset are summarized in \cref{tab-02}. Considering all 68 landmarks, all methods are relatively close for the paintings test set, except for the retrained version of dlib on art images that did not work well. By comparing only the inner facial landmarks (51) for which we apply the region refinement, the errors of all methods are considerably reduced, in particular for our ArtFacePoints. The facial landmark detection in prints is more challenging, resulting in overall higher errors. Our ArtFacePoints achieves the lowest error for both the total 68 and also the inner 51 landmarks. 
For both paintings and prints, fine-tuning of HR-Net shows improvements compared to the pretrained model, hence in the next experiments, we only include dlib (pretrained) and HR-Net (fine-tuned).

In our ablation study, we compare the global network only (trained for 60 epochs) and two versions of regional refinement: firstly, ArtFacePoints (w/o FM), for which we only use the RGB images as input to the region networks, \ie without the feature maps and secondly, our proposed ArtFacePoints which uses the concatenation of the RGB channels and the feature maps. We can observe for both paintings and prints that the regional refinement of both variants brings some benefit for the inner facial landmarks. Using the additional input of the feature maps is slightly superior, but ArtFacePoints (w/o FM) works a bit better for including the landmarks of the jaw line.

In \cref{fig-05}, we separately analyze the performance of the individual facial parts. Both variants of ArtFacePoints achieve the lowest errors for eyes, nose, mouth, and brows. The competing method Face of Art is relatively close to ours for the mouth, and it also works good for eyes and nose. HR-Net (fine-tuned) is the third best for the brows, for the eyes and mouth, HR-Net and dlib are comparable for the paintings, but HR-Net is better for the prints and also dlib did not detect landmark estimates for all faces. 
In general, the largest errors are obtained for the jaw line, which does not show as distinctive features as \eg the eyes and thus there is more ambiguity in the labeling process where to exactly position the landmarks on the face boundary and sometimes the face boundary is really hard to detect in case of occlusions by beard or hair. This uncertainty is hence also propagated into the models' predictions. The jaw line prediction of dlib is best for the paintings. As we used dlib as initial estimate for the manual labeling and then corrected the landmarks, there might be some bias for dlib regarding the jaw line. For the prints, dlib did not work so well, hence we also had a larger correction effort, which is also visible at the jaw line results. Our method has its limitations for the prediction of jaw lines, which where not specifically refined as only regions of the inner facial landmarks are extracted, but for instance for the registration of portraits only the landmarks of eyes, nose, and mouth are important, for which our method performs best.

Some visual examples for the landmark prediction results are shown in \cref{fig-06} for the paintings and in \cref{fig-07} for the prints. For each artwork, we additionally select a zoom in region for a precise comparison of the difference of the competing methods to our ArtFacePoints. For the paintings, in the first zoom region in \cref{fig-06}, the landmarks of the eye are most accurately predicted by ArtFacePoints, for dlib, HR-Net and Face of Art the eye is a bit too small. The second and third zoom regions in \cref{fig-06} depict the mouth region, for which ArtFacePoints most precisely detects the upper and lower lip. For the prints in \cref{fig-07}, dlib does not achieve an acceptable result for the first two images. In the zoomed regions of the eyes of all three examples, Face of Art and HR-Net miss the eye boundary in some corners and thus are less accurate than our ArtFacePoints.

Further, we tested the facial landmark detection for the low-resolution public Artistic Faces dataset (see \cref{tab-03}). As our method requires input resolution of $1024 \times 1024$, we upscale the $256 \times 256$ images to feed them to ArtFacePoints. The mean error is calculated at the low-resolution scale. Regarding the 68 and 51 facial landmarks Face of Art is slightly superior to our method, which is due to our lower performance of the jaw line prediction. However, for the individual regions of the core facial landmarks, ArtFacePoints is on par to Face of Art. HR-Net and dlib numerically also perform quite well for the low-resolution images, except that for dlib only 134 of 160 images could be considered to compute the ME due to dlib's false negative face detections. Thus, we could show that our ArtFacePoints can also be applied to the low-resolutions, but its advantage lies in the accurate prediction of the inner facial features for high-resolution applications.

\section{Applications}
In this section, we present some examples for the application of our ArtFacePoints to support the visual comparison of similar artworks.

\subsection{Image registration using facial landmarks}
To be able to visually compare two portraits based on their facial characteristics the images need to be registered. Therefore, the facial landmarks in both images serve as control point pairs for the registration. Since we want to explicitly align the eyes, nose, and mouth in the images, we only take these 41 landmarks as control points. 
Our aim is to find a global transformation matrix that transforms the source image in such as way that the error between the transformed source control points and target control points is minimal. To robustly compute the transform, we use random sample consensus (RANSAC)~\cite{FischlerMA1981} that estimates multiple transforms based on random subsets of the 41 control points and then selects the transform with the largest support based on all 41 control points.
The aspect ratio of the source image should be kept before and after the registration, thus we estimate a partial affine transform that includes rotation, translation, and scaling but no shearing. This is important for the similarity comparison of the faces to assess \eg if artists reused some facial structures in two different artworks of a similar motif.

\begin{figure}[t]
\centering
\includegraphics[width=\textwidth]{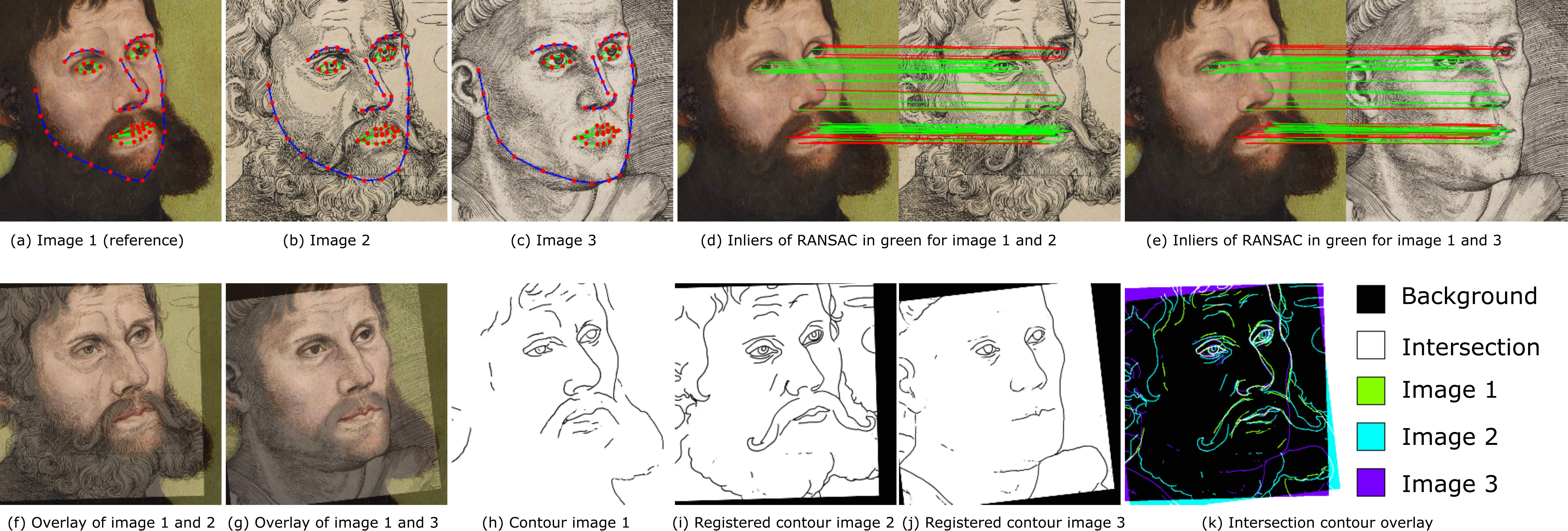}
\caption{Registration of paintings and prints using ArtFacePoints: Results for facial landmark detection, matching of inner facial landmarks, image overlays, contours generated using Art2Contour, and intersection contour overlay.\\ \\
\tiny{Image sources: Details of (a) Lucas Cranach the Elder, Martin Luther as ``Junker Jörg'' (Detail), Museum der bildenden Künste Leipzig;
(b) Lucas Cranach the Elder, Martin Luther as ``Junker Jörg'' (Detail, mirrored), Germanisches Nationalmuseum Nürnberg, Mp 14637;
(c) Lucas Cranach the Elder, Martin Luther as augustinian monk (Detail, mirrored), Klassik Stiftung Weimar, Museen, DK 182/83
}
}
\label{fig-08}
\end{figure}

\subsection{Facial image and contour comparison}
For the visual illustration of facial landmarks based registration in \cref{fig-08}, we picked three artworks by Cranach the Elder that depict Martin Luther as ``Junker Jörg'' (image 1 and 2) or augustinian monk (image 3). We detected the facial landmarks using ArtFacePoints and then applied RANSAC to predict the transformations between the reference image which is the painting and the two other images, the prints. For both registration pairs, in (d) and (e) the lines connecting the control points in the reference and source image are colored in green and red.
The green lines indicate the inliers, which are those points that were selected by RANSAC to compute the transformation and the red lines indicate the outliers that were excluded for the computation. Both examples show a high number of inliers. The first two images of the bottom row of \cref{fig-08} visualize the registration results as blended image overlays using alpha blending between the target and transformed source image, which indicate high similarity between the facial structures. 

Another possibility is to focus only on the facial contours for the comparison. For this task, we generate contour drawings for the three examples using the conditional generative adversarial network Art2Contour~\cite{SindelA2020} and apply the same transformations that we have predicted using the facial landmarks to warp the contour images correspondingly (see (h)-(j) in \cref{fig-08}). Then, we compute an intersection contour overlay in (k). It depicts the contours in white if at least two contours are intersected, and otherwise, \ie without any intersection, they are drawn in an image-specific color. That allows the comparison of multiple contours in one image, with a direct assignment of contour parts to the images. The intersection overlay depicted in \cref{fig-08} (k) for the given example images shows that these cross-modal pairs of painting and prints have a very similar shape of the main facial contour with some artistic differences in the contour line of the nose (painting) and in the right part of the jaw line (print with monk).

\section{Conclusions}
We presented a deep learning method for facial landmark detection in high-resolution artistic images. We employ a heatmap-based global network for coarse coordinates extraction and multiple heatmap-based region networks that operate on high image resolution only for specific regions. To train our method, we created a large synthetically augmented high-resolution dataset by using artistic style transfer and geometric transformations. In the experiments, we showed on our test dataset of paintings and prints, that our method, in comparison to competing methods, more accurately detects the facial landmarks of eyes, nose, and mouth, which are important for facial image registration. 
Then, we visually demonstrated for some example images the application of facial landmarks for cross-genre registration of paintings and prints and the possibility for facial image and contour comparison. Our method works precisely for the high-resolution landmarks of paintings and prints with moderate artistic shape and texture variations, but its performance of the detection of the jaw line is limited as we do not apply any refinement for this landmark group. Thus, future work will investigate model-based approaches to tune the detection of the facial outline.

\subsubsection{Acknowledgements} 
Thanks to Daniel Hess, Oliver Mack, Daniel G\"orres, Wibke Ottweiler, GNM, and Gunnar Heydenreich, CDA, and Thomas Klinke, \mbox{TH K\"oln}, and Amalie H\"ansch, FAU Erlangen-N\"urnberg for providing image data, and to Leibniz Society for funding the research project ``Critical Catalogue of Luther portraits (1519 - 1530)'' with grant agreement No. SAW-2018-GNM-3-KKLB, to the European Union’s Horizon 2020 research and innovation programme within the Odeuropa project under grant agreement No. 101004469 for funding this publication, and to NVIDIA for their GPU hardware donation. 

%\clearpage
% ---- Bibliography ----
%
% BibTeX users should specify bibliography style 'splncs04'.
% References will then be sorted and formatted in the correct style.
%
\bibliographystyle{splncs04}
\bibliography{ref}
\end{document}